\documentclass[10pt,twocolumn,letterpaper]{article}

\usepackage[preprint]{cvpr}      

\usepackage{graphicx}
\usepackage{amsmath}
\usepackage{amssymb}
\usepackage{booktabs}
\usepackage{wrapfig}  
\usepackage{color}
\usepackage{diagbox}
\usepackage{wrapfig}
\usepackage[ruled,linesnumbered]{algorithm2e}
\usepackage{makecell}  
\usepackage{amssymb}
\usepackage{pifont}
\usepackage{placeins}
\usepackage{selectp}

\newcommand{\D}{\mathcal{D}}

\newcommand{\R}{\mathbb{R}}

\newcommand{\w}{{\mathbf{w}}}
\newcommand{\m}{{\mathbf{m}}}
\usepackage[switch]{lineno}

\usepackage[pagebackref,breaklinks,colorlinks]{hyperref}

\usepackage{bbding}
\usepackage{colortbl}
\usepackage[capitalize]{cleveref}
\crefname{section}{Sec.}{Secs.}
\Crefname{section}{Section}{Sections}
\Crefname{table}{Table}{Tables}
\crefname{table}{Tab.}{Tabs.}

\def\cvprPaperID{3122} 
\def\confName{CVPR}
\def\confYear{2023}

\begin{document}

\title{OptG: Optimizing Gradient-driven Criteria in Network Sparsity}

\author{Yuxin Zhang$^1$ \quad Mingbao Lin$^{2}$ \quad Mengzhao Chen$^{1}$ \quad Fei Chao$^1$ \ \quad Rongrong Ji$^{1,3,4}$\thanks{Corresponding author: rrji@xmu.edu.cn}
  \\[0.2cm] 
  $^1$Media Analytics and Computing Laboratory, Department of Artificial Intelligence,\\
  School of Informatics, Xiamen University, Xiamen, China\\
  $^2$Tencent Youtu Lab, Shanghai, China\\
  $^3$Institute of Artificial Intelligence, Xiamen University, Xiamen China \\
  $^4$Pengcheng Lab, Shenzhen, China
}

\maketitle

\begin{abstract}
Network sparsity receives popularity mostly due to its capability to reduce the complexity of the network. Extensive studies excavate gradient-driven sparsity. Typically, these methods are constructed upon the premise of weight independence, which however, is contrary to the fact that weights are mutually influenced. Thus, their performance remains to be improved.
In this paper, we propose to optimize gradient-driven sparsity (OptG) by solving this independence paradox. 
Our motive comes from the recent advances in supermask training which shows that high-performing sparse subnetworks can be located by simply updating mask values without modifying any weight. 
We prove that supermask training is to accumulate the criteria of gradient-driven sparsity for both removed and preserved weights, and it can partly solve the independence paradox. 
Consequently, OptG integrates supermask training into gradient-driven sparsity, and a novel supermask optimizer is further proposed to comprehensively mitigate the independence paradox.
Experiments show that OptG can well surpass many existing state-of-the-art competitors, especially at ultra-high sparsity levels.
Our project is available at \url{https://github.com/zyxxmu/OptG}.
\end{abstract}

\section{Introduction}
\label{sec:intro}
The explosive advances of convolution neural networks (CNNs) are mainly driven by continuously growing model parameters, incurring deployment difficulty on resource-constrained devices.
By directly removing parameters for a sparse model, network sparsity emerges as an important technique to reduce model complexity~\cite{lecun1989optimal,mozer1989skeletonization,hoefler2021sparsity}.
Broadly speaking, methods in the literature can be divided into after-training sparsity, before-training sparsity and during-training sparsity~\cite{liu2021sparse}. After-training sparsity aims to remove parameters in pre-trained models~\cite{han2015learning}, while before-training sparsity attempts to refrain from the time-consuming pre-training process by constructing sparse models at random initialization~\cite{lee2018snip}. Recent advances advocate during-training sparsity which consults the sparsity process throughout network training~\cite{evci2020rigging, kusupati2020soft}.
Particularly, the core of network sparsity lies in selecting to-be-removed parameters such that it can satisfy: 1) desired sparse rate; 2) acceptable performance compromise. 
To this end, the most straightforward solution is to remove these parameters causing the least increase in the training loss $\mathcal{L}$. Then, by leveraging the first-order Taylor expansion of the loss function $\mathcal{L}$ to approximate the influence of removing parameter $\w_i$, the key criteria for measuring weight importance can be expressed as $\frac{\partial \mathcal{L}}{\partial \w_i}\w_i$, leading to gradient-driven sparsity. Recent advances rewrite this criteria using higher-order Taylor expansion~\cite{wang2020picking, molchanov2016pruning}, which will be detailed in the next section.

\begin{figure}[!t]
\begin{center}
\includegraphics[height=0.63\linewidth]{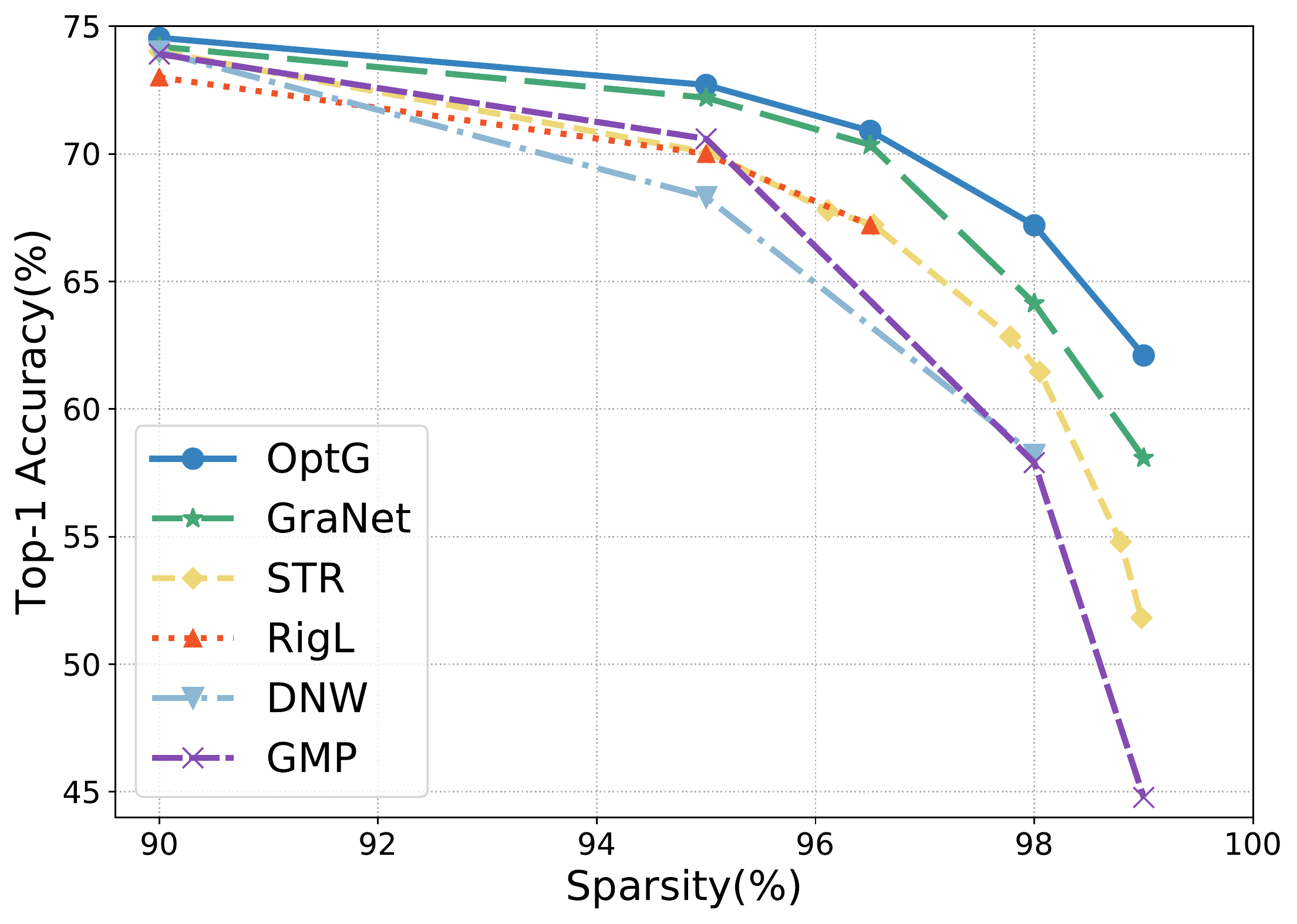}
\end{center}
\centering\vspace{-1.5em}
\caption{\label{fig:acc}Top-1 accuracy \emph{v.s.} sparsity with ResNet-50 on ImageNet. The proposed OptG significantly improves performance of other approaches, especially at the ultra-high sparsity levels.
}
\vspace{-1.3em}
\end{figure}

Despite the progress, existing gradient-driven methods are built upon the premise of independence among weights. However, this assumption contradicts with the practical implementation, in which, parameters are collectively making effort to derive the network output. Usually, existing methods remove weights once-for-all~\cite{lee2018snip, wang2020picking}. Consequently, the computed loss change used to remove weights deviates a lot from the actual loss change and such deviation is proportional to the number of removed weights at one time. Thus, it is necessary to overcome this independence paradox in order to pursue a better performance.

%

Beyond the gradient-driven sparsity, recent developments on supermask training~\cite{zhou2019deconstructing, ramanujan2020s, zhang2021lottery} show that high-performing sparse subnetworks can be located without modifying any weight. Instead, they choose to update mask values using the straight-through-estimator (STE)~\cite{bengio2013estimating}. In this paper, we innovatively prove that the essence of supermask training is to accumulate the criteria in gradient-driven sparsity for both preserved and removed weights. Also, we show that this manner can partially solve the independence paradox. Unfortunately, the fixed weights in supermask training fail to eliminate the independence paradox, thus, the performance is still sub-optimal.

In this paper, we propose to optimize the gradient-driven sparsity by integrating the advantage of supermask training in overcoming the independence paradox. 
Our method, termed OptG, conducts sparse weights and supermask training simultaneously with a novel supermask optimizer that continuously accumulates the mask gradients of each training iteration and only updates the mask at the beginning of each training epoch. In this way, the remaining parameters can be well tuned on the training set to eliminate the error gap.
We further equip the supermask optimizer with a sparsity-aware learning rate schedule that allocates the learning rate of supermasks proportional to the sparsity level of current training iteration, such that the deviation caused by the independence paradox can be further reduced, enabling an optimization for the gradient-driven sparsity.
Extensive experiments demonstrate that OptG dominates its counterparts, especially at extreme sparse rate that suffers most significant error gap from the independence paradox.
For instance, OptG removes 98\% parameters of ResNet-50 while still achieving 67.20\% top-1 accuracy on ImageNet, surpassing the recent strong baseline STR~\cite{kusupati2020soft} that only reaches 62.84\% by a large margin.

%

Our main contributions are summarized below:
\begin{itemize}
    \item We prove the existence of the independence paradox in gradient-driven sparsity, which causes an error gap in loss change for measuring weight importance.
    
    \item We reveal the essence of supermask training is to accumulate weights gradients in gradient-driven sparsity, which partly solves the independence paradox.
    
    \item We propose OptG that further optimizes the gradient-driven sparsity using a novel mask optimizer that overcomes the problem of the independence paradox.
    
    \item Extensive experiments demonstrate the advantage of the proposed OptG over many existing state-of-the-arts in sparsifying modern CNNs.
\end{itemize}

\section{Related Work}
This section covers the spectrum of studies on sparsifying CNNs that closely related to our work. A more comprehensive overview can be found in the recent survey~\cite{hoefler2021sparsity}.

\subsection{Sparsity Granularity}
The granularity of network sparsity varies from coarse grain to fine grain.
The former is indicated to removing the entire channels or filters towards a structured subnetwork~\cite{he2017channel, lin2020hrank, he2019filter}. Though well suited to a practical speedup on regular hardware devices, significant performance degradation usually occurs at a high sparse rate~\cite{ding2018auto,luo2017thinet,he2020learning, he2018soft}.
The latter removes individual neurons at any location of the network to pursue an unstructured subnetwork~\cite{han2015learning, mocanu2018scalable}.
It has been proved to well retain the performance even under an extremely high sparse rate~\cite{kusupati2020soft, gale2019state}.
Moreover, many recent efforts also show great promise of unstructured sparse networks in practical acceleration~\cite{gale2020sparse,elsen2020fast,zhou2021learning,zhang2022learning}.
In particular, the recent 2:4 sparse pattern has been well supported by Nvidia A100 GPUs to accomplish 2$\times$ speedups.

%
%
%

\subsection{When to Sparsify}
According to the time point the sparsity is applied, we empirically categorize existing works into three groups~\cite{liu2021sparse}. 

\textbf{After-training sparsity} was firstly adopted by the optimal brain damage~\cite{lecun1989optimal}. Since then, the followers obey a three-step pipeline, including model pretraining, parameter removing and network fine-tuning~\cite{han2015learning,molchanov2016pruning,ding2019centripetal, lemaire2019structured}. 
%
%
Unfortunately, in the cases that pre-trained models are missing and hardware resources are limited, the aforementioned approaches become impractical due to the expensive fine-tuning process.

\textbf{Before-training sparsity} attempts to conduct network sparsity on randomly initialized networks for efficient model deployment. Trough removing weights using gradient-driven measurement~\cite{lee2018snip, wang2020picking} or heuristic design~\cite{tanaka2020pruning}, a sparse subnetwork can be available in a one-shot manner.
Nevertheless, the performance gap of this group still exists compared with the after-training sparsity~\cite{frankle2020pruning}.

\textbf{During-training sparsity} has been drawing increasing attention for its performance retaining~\cite{mocanu2018scalable,mostafa2019parameter,lin2020dynamic}.
In each training iteration, the parameters will be removed or revived according to a predefined criterion. Consequently, the sparse subnetwork can be obtained in a single training process.
For instance, RigL~\cite{evci2020rigging} re-allocates the removed weights according to their dense gradients, while Sparse Momentum~\cite{dettmers2019sparse} considers
the mean momentum magnitude of each layer as a redistribution criterion. Besides, the performance of during-training sparsity can be further enhanced if the desired sparsity is gradually achieved in an incremental manner~\cite{zhu2017prune,liu2021sparse,liu2021we}.

\subsection{Layer-wise Sparsity Allocation}
It has been a wide consensus in the community that layer-wise sparsity allocation, \emph{i.e.}, sparse rate of each layer, is a core in network sparsity~\cite{liu2018rethinking, gale2019state, lee2020layer}. Majorities of existing methods implement layer-wise sparsity using a static or dynamic design~\cite{evci2020rigging,lee2020layer}.
Typically, the global sorting manner also leads to a non-uniform sparsity allocation~\cite{han2015learning}.
Unfortunately, as pointed by~\cite{tanaka2020pruning}, this may result in an extremely imbalanced sparsity budget that parameters in some layers are mostly removed, which further disables the network training.
Therefore, recent studies~\cite{kusupati2020soft,savarese2019winning} pursue the layer-wise sparsity in a trainable manner, which, however, requires a complex hyper-parameter tunning and often leads to an unstable sparse rate.
%
%
Different from the above approaches, our OptG can safely conduct pruning in a global manner to automatically obtain the layer-wise sparsity allocation, without complex parameter tuning.

\subsection{Lottery Ticket Hypothesis and the Supermask}
The lottery ticket hypothesis~\cite{frankle2018lottery} reveals that there exist randomly-initialized sparse networks that can be trained independently to match the performance of the dense model.
Following this conjecture, recent empirical studies~\cite{zhou2019deconstructing,ramanujan2020s, zhang2021lottery} have further confirmed the existence of supermask, which simply updates the mask values to obtain sparse subnetworks using the straight-through-estimator (STE)~\cite{bengio2013estimating} without the necessity of modifying weight values.
For instance, ~\cite{ramanujan2020s} showed that a randomly initialized Wide ResNet-50 sparsified by a supermask, can match the performance of ResNet-32 trained on ImageNet. \cite{orseau2020logarithmic,pensia2020optimal} further proved that the existence of supermask relies on a logarithmic over-parameterization.
Despite the progress, an in-depth analysis remains unexplored on why a subnet exists without modifying weight values.

%

\section{Methodology}

\subsection{Background}\label{background}
\textbf{Notations}. Denoting the weights of a convolution neural network as $\w \in \R^N$ where $N$ is the weight number, network sparsity can be viewed as multiplying a binary mask $\m \in \{0, 1 \}^N$ on $\w$ as ($\w \odot \m $) where $\odot$ represents the element-wise product.
Consequently, the state of the $i$-th mask $\m_i$ indicates whether $\w_i$ is removed ($0$) or not ($1$).

Let $\mathcal{L}(\cdot)$ and $\D$ be the training loss and training dataset.
Essentially, given a sparse rate $P$, network sparsity aims to obtain a sparse $\mathbf{m}$ subject to $\frac{{||\m||}_0}{N}  \leq 1 - P$, meanwhile minimizing $\mathcal{L}(\cdot)$ on $\D$.
To this end, various scenarios have been proposed to derive $\m$. In what follows, we discuss two cases that are mostly related to our method.

\textbf{Gradient-driven sparsity}.
The studies of network sparsity using weight gradient date back to the last few decades~\cite{lecun1989optimal, mozer1989skeletonization}. The basic idea of these methods is to leverage weight gradients to approximate change in the loss function $\mathcal{L}(\cdot)$ when removing some parameters. The overall optimization can be formulated as:
\begin{equation}
 \min_{\w} \mathcal{L}(\w \odot \m\; ; \;\D) \;\;\;
\emph{s.t.} \;\;\; \frac{{||\m||}_0}{N}  \leq 1 - P.
  \label{eq1}
\end{equation}

Then, it is easy to know that the loss change after removing a single weight $\w_i$ is:
\begin{equation}
  \Delta \mathcal{L}(\w_i ; \D) =   \mathcal{L}(\m_i=0 ; \D)-\mathcal{L}(\m_i = 1 ; \D).
\end{equation}

It is intuitive that $\Delta \mathcal{L}(\w_i ; \D) < 0$ indicates a loss drop, which means the removal of $\w_i$ results in better performance. To obtain a sparse $\m$, one naive approach is to repeatedly compute the loss change for each weight in $\w$, and then set masks of these parameters with smaller $\Delta \mathcal{L}(\w_i ; \D)$ to $0$s, and $1$s otherwise.
However, modern CNNs tend to have parameters in millions, making it expensive to perform this one-by-one loss calculation.
Fortunately, $\Delta \mathcal{L}(\w_i ; \D)$ can be approximated via the Taylor series expansion. Considering the first-order case~\cite{molchanov2016pruning}, $\Delta \mathcal{L}(\w_i ; \D)$ can be reformulated as:
\begin{equation}
\begin{split}
\Delta \mathcal{L}&(\w_i ; \D) =  \mathcal{L}(\m_i=0 ; \D)-\mathcal{L}(\m_i = 1 ; \D) \\
& = \mathcal{L}(\m_i=1 ; \D)- \frac{\partial \mathcal{L}}{\partial (\w_i \odot \m_i)} (\w_i \odot \m_i) \\
& + R_1(\m_i=0) -  \mathcal{L}(\m_i = 1 ; \D) \\
& =  -\frac{\partial \mathcal{L}}{\partial (\w_i \odot \m_i)} (\w_i \odot \m_i)  + R_1(\m_i=0).
\end{split}
\label{eq2}
\end{equation}

If we ignore the first-order remainder $R_1(\m_i=0)$,  then:
\begin{equation}
\Delta  \mathcal{L}(\w_i ; \D) \approx -\frac{\partial \mathcal{L}}{\partial (\w_i \odot \m_i)} (\w_i \odot \m_i).
\label{noremainder}
\end{equation}

Eq.\,(\ref{noremainder}) can be an efficient alternative to approximating $\mathcal{L}(\w_i ; \D)$, since for all weights, the term $\w_i \odot \m_i$ can be available in a single forward propagation and the term $-\frac{\partial \mathcal{L}}{\partial (\w_i \odot \m_i)}$ can be derived in a single backward propagation. Consequently, the format of Eq.\,(\ref{noremainder}) has served as a basis in modern gradient-driven network sparsity~\cite{mozer1989skeletonization}. 
Many recent variants are further excavated based on this format.
Taylor-FO~\cite{molchanov2016pruning} considers $(\frac{\partial \mathcal{L}}{\partial \w} \odot \w) ^ 2$ as a saliency metric to sparsify a pre-trained model, which is similar to the prune-at-initialization SNIP~\cite{lee2018snip} that uses $|\frac{\partial \mathcal{L}}{\partial \w} \odot \w|$ instead.
GrasP~\cite{wang2020picking} leverages the second-order Taylor series and derives the pruning criterion of $-\mathbf{H}\frac{\partial \mathcal{L}}{\partial \w} \w$, where $\mathbf{H}$ denotes the Hessian matrix.
Besides, another variant $|\frac{\partial \mathcal{L}}{\partial (\w_i \odot \m_i)}|$ is used to indicate if some pruned weights should be revived during sparse training.

Though great effort has been made, these existing works are developed on the premise of independence assumption that weights are irrelevant to each other, which however, is on the contrary in practice. As results, their performance remains an open issue.

\textbf{Supermask-driven sparsity}.
In gradient-driven sparsity, the values of weight vector $\w$ are updated in the backward propagation and the mask $\m$ is recomputed using the above criterion in the forward propagation. Instead, many recent developments reveal that a high-performing sparse subnet can be found without the necessity of modifying any weight~\cite{zhou2019deconstructing,zhang2021lottery,zhou2019deconstructing}. Typically, these methods can be complemented by updating the mask vector $\m$. The corresponding learning objective can be formulated as:
\vspace{-0.5em}
\begin{equation}
 \min_{\m} \mathcal{L}(\w \odot \m\; ; \;\D) \;\;\;
\emph{s.t.} \;\;\; \frac{{||\m||}_0}{N}  \leq 1 - P.
  \label{eq4}
\end{equation}

To stress, the objective of Eq.\,(\ref{eq4}) differs from that of Eq.\,(\ref{eq1}) in that its optimized variable is $\m$ instead of $\w$ which instead is regarded as a constant vector in Eq.\,(\ref{eq4}). Existing studies~\cite{zhou2019deconstructing,ramanujan2020s} optimize Eq.\,(\ref{eq4}) by first relaxing the discrete $\m \in \{0,1\}^N$ to a continuous version of $\hat{\m} \in \R^N$. Then, in the forward propagation, the discrete mask $\m$ is generated by applying a binary function $h(\cdot)$ to $\hat{\m} \in \R^N$ as:
\vspace{-0.5em}
\begin{equation}
h(\mathbf{\hat{m}}_i) = \left\{ \begin{array}{ll} 
 0, \textrm{if $\mathbf{\hat{m}}_i$ in the top-$\lceil P \cdot N \rceil$ smallest values of $\mathbf{\hat{m}}$,}\\
 1, \textrm{otherwise.}
  \end{array} \right.
\label{eq5}
\end{equation}

In the backward propagation, due to the non-differentiable in the above equation, the straight-through estimator (STE)~\cite{bengio2013estimating} is used as an alternative to approximate the mask gradient as:
\vspace{-0.5em}
\begin{equation}
\begin{split}
    \frac{\partial \mathcal{L}}{\partial \mathbf{\hat{m}}_i} & = \frac{\mathcal{\partial L}}{\partial 
    (h(\mathbf{\hat{m}}_i)\odot\w_i)}\frac{\partial (h(\mathbf{\hat{m}}_i)\odot\w_i)}{\partial h(\mathbf{\hat{m}}_i)}\frac{h(\mathbf{\hat{m}}_i)}{\mathbf{\hat{m}}_i} \\
    & \approx \frac{\mathcal{\partial L}}{\partial(h(\mathbf{\hat{m}}_i)\odot\w_i)}\frac{\partial(h(\mathbf{\hat{m}}_i)\odot\w_i)}{\partial h(\hat{\m}_i)} 
    \\&
    =\frac{\mathcal{\partial L}}{\partial(h(\mathbf{\hat{m}}_i)\odot\w_i)}\w_i.
\end{split}
 \label{mask_gradient}
\end{equation}

By updating $\hat{\m}$, a high-performing sparse subnet can be finally located. Nevertheless, to date, no one has dived into an exploration of how a subnet can be identified without modifying weight values. In Sec.~\ref{insights}, we give a detailed explanation and show that gradient-driven sparsity and supermask-driven sparsity are the same in essence.

\subsection{Independence Paradox\label{paradox}}
The gradient-driven sparsity using Eq.\,(\ref{noremainder}) neglects the high-order terms of Taylor series as well as the remainder. Luckily, it has been experimentally proved that the first-order gradient~\cite{lee2018snip} shows performance on par with the higher-order ones~\cite{wang2020picking}.
Nevertheless, existing gradient-driven sparsity is built upon the assumption that weights are irrelevant to each other, which is contrary to the practical implementation~\cite{molchanov2016pruning, lee2018snip} where a large number of weights are usually removed simultaneously. Considering the case that two weights $\w_i$ and $\w_j$ are removed, if independent, the loss change of removing $\w_i$ using Eq.~(\ref{eq2}) is:
\begin{equation}
\begin{split}
\Delta   \mathcal{L}(\w_i ; \D) &=  \mathcal{L}(\m_i=0, \m_j=1, \w; \D) \\
&-\mathcal{L}(\m_i = 1, \m_j = 1, \w ; \D). 
\label{eq7}
\end{split}
\end{equation}
However, considering that $\w_j$ has been removed as well, the actual loss change due to the removal of $\w_i$ should become:
\begin{equation}
\begin{split}
\Delta\mathcal{L}(\w_i^* ; \D) &=  \mathcal{L}(\m_i=0, \m_j=0, \w^* ; \D)\\ &-\mathcal{L}(\m_i = 1, \m_j = 0, \w^* ; \D),
\label{eq8}
\end{split}
\end{equation}
where $\w^*$ indicates the state of original $\w$ after the removal of $\w_j$ and a follow-up fine-tuning on the dataset $\mathcal{D}$.
It is easy to know that Eq.\,(\ref{eq7}) is actually built upon the premise of preserving $\w_j$. However, the practice removes $\w_i$ and $\w_j$ simultaneously, which indicates a loss change of Eq.\,(\ref{eq8}). 
As tremendous weights are removed at once in the real cases~\cite{lee2018snip, wang2020picking}, such variants of loss change between Eq.~(\ref{eq7}) and Eq.~(\ref{eq8}) could raise sharply, which we quantitatively demonstrate in the supplementary material.
Thus, there exists an independence paradox in existing studies and the error gap in gradient-driven sparsity is proportional to the total number of removed weights each time. 
%

%
%
%

Note that, some recent advances~\cite{evci2020rigging,zhu2017prune} advocate incremental pruning that removes a small portion of weights each time. 
For instance, RigL~\cite{evci2020rigging} removes a small fraction of weights and activates new ones iteratively, while Zhu~\emph{et al.}~\cite{zhu2017prune} proposed to gradually increase the number of removed weights until the desired sparse rate is satisfied.
Though not explicitly stated, these works indeed accomplish network sparsity by reducing the removed parameters so as to relieve the error gap caused by the independence paradox. 
Nevertheless, the error gap still exists and thus their performance is sub-optimal.
Therefore, an in-depth exploration to overcome this independence paradox remains to be done.

\subsection{Our Insights on Supermask-driven Sparsity}\label{insights}
In this subsection, we show that the success of supermask-driven sparsity, to some extent, mitigates the aforementioned independence paradox.
We first prove that the mechanism of supermask-driven sparsity actually meets the first-order based gradient-driven sparsity. 
Specifically, let the mask $\hat{\m}_i$ at the $t$-th training iteration be $\hat{\m}_i^t$. Combining the mask gradient in Eq.\,(\ref{mask_gradient}), $\hat{\m}_i^t$ can be derived via SGD as:
\vspace{-0.5em}
\begin{equation}
\begin{split}
    \hat{\m}_i^t &= \hat{\m}_i^{t-1} -   \eta \frac{\partial \mathcal{L}}{\partial \mathbf{\hat{m}}_i^t}\\ &= \hat{\m}_i^{t-1} -  \eta \frac{\partial \mathcal{L}}{ \partial(h(\mathbf{\hat{m}}_i^t)\odot\w_i)} \w_i, 
\label{mask_tth}
\end{split}
\end{equation}
where $\eta$ indicates the learning rate. Note that the momentum and weight decay items are neglected here for simplicity. 
When $h(\mathbf{\hat{m}}^t_i) = 1$, we have $(h(\mathbf{\hat{m}}_i^t)\odot\w_i) = \w_i$, which leads the updating items of $\hat{\m}_i^t$ to the result of Eq.~(\ref{noremainder}). 
Nevertheless, when $h(\mathbf{\hat{m}}_i) = 0$, which indicates a removal of $\w_i$, the computing result of Eq.\,(\ref{mask_tth}) remains unclear. 
Here we show that in this case, the updating items of $\hat{\m}_i^t$ become the loss change for adding back the corresponding pruned weight $\w_i$, which can be derived as:
\vspace{-0.3em}
\begin{equation}
\begin{split}
\Delta^{+}  & \mathcal{L}(\w_i ; \D) =  \mathcal{L}(\m_i=1 ; \D)-\mathcal{L}(\m_i = 0 ; \D) \\
& = \mathcal{L}(\m_i=0 ; \D)- \frac{\partial \mathcal{L}}{\partial (h(\mathbf{\hat{m}}_i^t)\odot\w_i)} (0 - \w_i) \\
& \quad+ R_1(\m_i=1) -  \mathcal{L}(\m_i = 0 ; \D) \\
& =  \frac{\partial \mathcal{L}}{\partial (h(\mathbf{\hat{m}}_i^t)\odot\w_i)} (h(\mathbf{\hat{m}}_i^t)\odot\w_i)  + R_1(\m_i=1) \\
& \approx  \frac{\partial \mathcal{L}}{\partial (h(\mathbf{\hat{m}}_i^t)\odot\w_i)} (h(\mathbf{\hat{m}}_i^t)\odot\w_i) \\
&= \frac{\partial \mathcal{L}}{ \partial(h(\mathbf{\hat{m}}_i^t)\odot\w_i)} \w_i.
\end{split}
\label{add_back}
\end{equation}

This is exactly the pursued mask gradient in Eq.\,(\ref{mask_tth}). Thus, the updating rule of $\hat{\m}_i$ can be organized as:

\vspace{-1em}
\begin{equation}
\hat{\m}_i^t = \left\{ \begin{array}{ll} 
 \hat{\m}_i^{t-1} + \eta \Delta   \mathcal{L}(\w_i ; \D) , \; \textrm{if $h(\mathbf{\hat{m}}^{t-1}_i) = 1$,}\\
 \hat{\m}_i^{t-1} -\eta \Delta^{+}\mathcal{L}(\w_i ; \D), \; \textrm{otherwise.}
  \end{array} \right.
\label{essence}
\end{equation}

As defined previously, $\Delta \mathcal{L}(\w_i ; \D)$ is the loss change after removing $\w_i$ and we expect a large value of $\Delta \mathcal{L}(\w_i ; \D)$ if $\w_i$ is important to the network performance.
Similarly, a small $\Delta^{+}\mathcal{L}(\w_i ; \D)$ also indicates the removed weight is vital to the network and should be revived in the case of supermask-driven sparsity.
Overall, from Eq.\,(\ref{essence}), we can know that the updating in supermask-driven sparsity is indeed to accumulate the criteria of both preserved and removed weights in gradient-driven sparsity. Thus, we can conclude that the manner of supermask-driven sparsity to obtain a sparse mask $\m$ is indeed similar to that of gradient-driven sparsity. 
Further, we show that the key to supermask training is that it partially solves the independence paradox. Similarly, given that $\w_i$ and $\w_j$ are removed at the ($t-1$)-th training iteration, we have $h(\hat{\m}_i^{t-1}) = 0$. According to Eq.\,(\ref{essence}), the loss change for adding back $\w_i$ becomes:
\vspace{0em}
\begin{equation}
\begin{split}
\Delta^{+}  \mathcal{L}(\w_i ; \D) &=  \mathcal{L}(\m_i=1, \m_j=0, \w ; \D)\\
&-\mathcal{L}(\m_i = 0, \m_j = 0, \w ; \D).
\end{split}
\label{closer}
\end{equation}

Obviously, Eq.\,(\ref{closer}) is closer to the actual loss change of Eq.\,(\ref{eq8}) than independent-based Eq.~(\ref{eq7}). 
Thus, the error gap from the independence paradox can be well compensated by reviving $\w_i$ if it is important to the network performance during the following training iterations.
This well explains why supermask-driven sparsity can perform well in existing studies~\cite{ramanujan2020s, zhang2021lottery}.
Nevertheless, as the weights are kept fixed during the supermask training, there is still a distance between Eq.\,(\ref{eq8}) and Eq.\,(\ref{closer}), which implies that the independence paradox is still not comprehensively solved yet.

\subsection{Optimizing Gradient-driven Sparsity}
Herein, we propose to interleave the supermask training into during-training learning to further optimize the gradient-driven sparsity. The learning objective of our method, termed OptG, can be formulated as:

\vspace{-1.2em}
\begin{equation}
 \min_{\m, \w} \mathcal{L}(\w \odot \m\; ; \;\D) \;\;\;
\emph{s.t.} \;\;\; \frac{{||\m||}_0}{N}  \leq 1 - P .
  \label{object_optg}
\end{equation}

The motive of our OptG is to conduct weight optimization and mask training simultaneously.
Therefore, when $\w_j$ is pruned, the remaining weights are further trained on $\D$ and the update item of $\hat{\mathbf{m}}_i$ in the $t$-th training iteration is:

\begin{figure*}[!t]
\begin{center}
\includegraphics[height=0.28\linewidth]{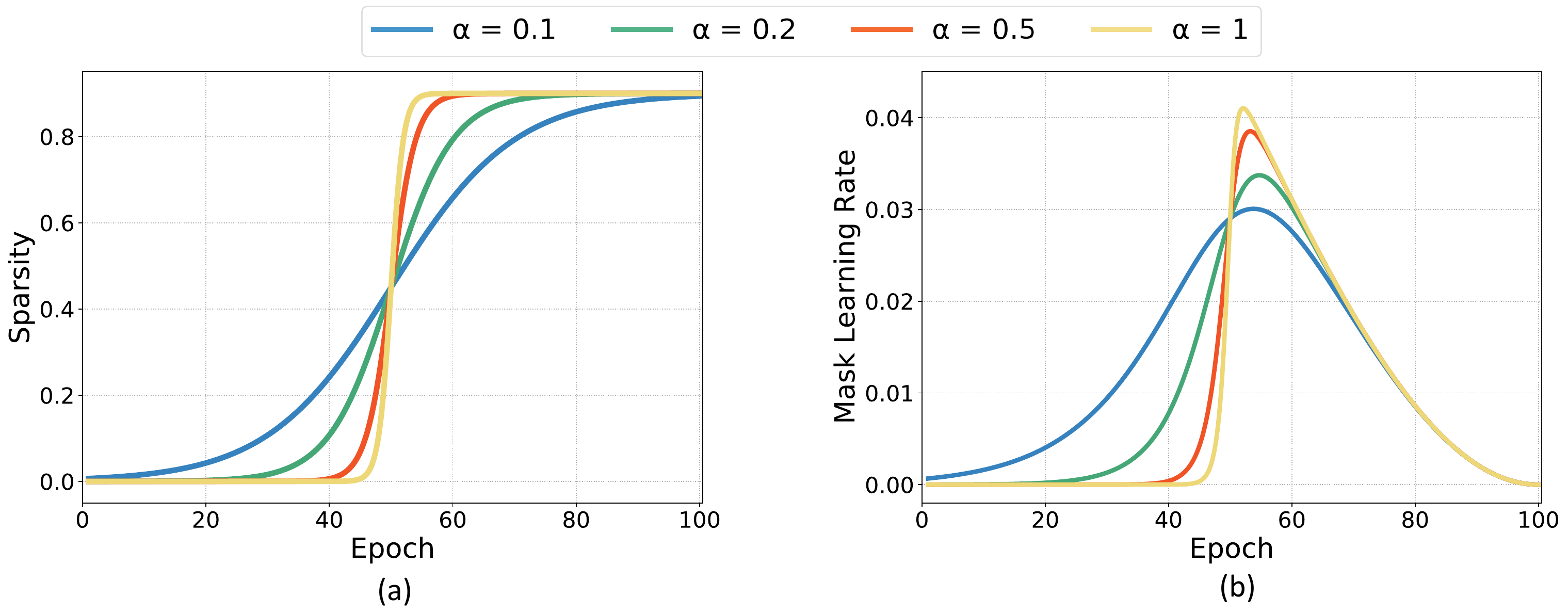}
\end{center}
\vspace{-1.5em}
\caption{\label{fig:mask_optimizer}(a) The progression of the overall sparsity and (b) the learning rate of the supermasks with different $\alpha$ over the course of training.
}
\vspace{-1em}
\end{figure*}

\vspace{-1.2em}
\begin{equation}
\begin{split}
\Delta   \mathcal{L}(\w_i ; \D) &=  \mathcal{L}(\m_i=1, \m_j=0, \w^t ; \D)\\
&-\mathcal{L}(\m_i = 0, \m_j = 0, \w^t ; \D), 
\label{eq14}
\end{split}
\end{equation}
where $\w^t$ is the trained weights after the $t$-th training iteration.
Nevertheless, it is clear that $\w^t$ barely reflects the weight tuning on the whole training set as $\w^t$ has been trained only for one iteration.
Moreover, if the binary function $h(\cdot)$, \emph{i.e.}, Eq.~(\ref{eq5}), is applied during each forward propagation of masks, the network topology may be changed frequently, which can lead to an unstable training process.
To solve the above-mentioned problem, we introduce a novel supermask optimizer towards comprehensively solving the independence paradox.
In particular, we apply $h(\cdot)$ to revive and prune weights at the beginning of each training epoch.
Then, we continuously accumulate the mask gradient during each training iteration via Eq.~(\ref{mask_tth}), but keep the binary mask fixed.
Therefore, the preserved weights can be sufficiently retrained on the training set, enabling our mask updating process to reach Eq.~(\ref{eq8}).
From another perspective, as discussed in Sec.~\ref{paradox}, the error caused by the independence paradox is actually proportional to the number of removed weights and an alternatively way to eliminate it is to gradually remove weights.
Nevertheless, existing studies~\cite{zhu2017prune, liu2021sparse} choose to rapidly improve the sparsity level during the early stage of training, which we point out does not fit our goal for optimizing gradient-driven sparsity.
To explain, previous techniques~\cite{zhu2017prune, liu2021sparse} revive weights to 0s, where our supermask training revive weights to the values before they are sparsified in order to mitigate the independence paradox, which implies that the weights require sufficient training as reviving a random-initialized weight is usually meaningless from the perspective of Eq.~(\ref{add_back}).
Thus, we choose to increase the sparse rate from 0 to the target sparsity rate $P$ in a more smooth manner as:
\begin{equation}
    P_k = \frac{P}{1+e^{-\alpha (k-0.5\tau)}},
    \label{eq15}
\end{equation}
where $k$ and $\tau$ represent the current and total training epoch, $\alpha$ is a hyperparameter that controls the total epoch for achieving sparsity.
Fig.~\ref{fig:mask_optimizer} (a) shows that the sparsity ascent rate at initialization can be relatively smooth and we experimentally proved in Sec.~\ref{ablation} that such sparsity schedule well boosts the performance of OptG for reviving weights to their original values.

Based on this schedule, we further embed a novel paradox-aware supermask learning rate schedule with our supermask optimizer. In detail, the mask learning rate at epoch $k$, denoted as $\eta_{\hat{\m},k}$, is calculated as: 
\begin{equation}
    \eta_{\hat{\m},k} =  \frac{\eta_{\w, k}}{1+e^{-\alpha (k-0.5\tau)}},
    \label{eq16}
\end{equation}
where $\eta_{\w, k}$ is the learning rate of weights at the $k$-th epoch scheduled by the cosine annealing~\cite{loshchilov2016sgdr}.
As can be reffered in Fig.~\ref{fig:mask_optimizer} (b), when the sparse rate of the network is low, the learning rate of masks is also small as the calculation of gradient score will be seriously interfered by the independence paradox. 
Moreover, when the sparse rate reaches $P$, the learning rate of masks can follow weights to sufficiently optimize the gradient-driven criteria while guaranteening training convergence.
We summarize the workflow of OptG in Alg.~\ref{alg:optg}. 
Note that our OptG sorts the weights in a global manner to automatically decide a layer-wise sparsity budget, thus avoiding the rule-of-thumb design~\cite{evci2020rigging} or complex hyper-parameter tuning for learning sparsity distributions~\cite{he2018soft}.

\section{Experiments}\label{experiment}

\subsection{Settings}
We conduct extensive experiments to evaluate the efficacy of our OptG in sparsifying VGGNet-19~\cite{simonyan2015very} on small scale CIFAR-10/100~\cite{krizhevsky2009learning} and ResNet-50~\cite{he2016deep}, MobileNet-V1~\cite{howard2017mobilenets} on large scale ImageNet~\cite{deng2009imagenet}.
Besides, we compare our OptG with several the state-of-the-arts including SNIP~\cite{lee2018snip}, GraSP~\cite{wang2020picking}, SET~\cite{mocanu2018scalable}, GMP~\cite{gale2019state},  SynFlow~\cite{tanaka2020pruning}, DNW~\cite{wortsman2019discovering}, RigL~\cite{evci2020rigging}, GSM~\cite{ding2019global}, STR~\cite{kusupati2020soft} and GraNet~\cite{liu2021sparse}.

We implement OptG with PyTorch~\cite{pytorch2015}.
Particularly, we set $\alpha = 0.5$ in all experiments and leverage the SGD optimizer to update the weights and their masks with a gradually-increasing sparsity rate Eq.\,(\ref{eq15}).
On CIFAR-10 and CIFAR-100, we train the networks for 160 epochs with a weight decay of $1\times10^{-3}$.
On ImageNet, the weight decay is set to $5\times10^{-4}$ for ResNet-50 and $4\times10^{-5}$ for MobileNet-V1.
We train ResNet-50 for 100 epochs and MobileNet-V1 for 180 epochs, respectively.
Besides, the initial learning rate is set to 0.1, which is then decayed by the cosine annealing scheduler~\cite{loshchilov2016sgdr}.
All experiments are run with NVIDIA Tesla V100 GPUs.

\begin{algorithm}[!t]
\SetKwInOut{Input}{Require}
\SetKwInOut{Output}{Output}
\caption{Optimizing the gradient-driven sparsity.}
\label{alg:optg}
\Input{ Network weights $\w$ and masks $\hat{\m}$, target sparsity $P$, total training epoch $\tau$.}
$\w$ $\gets$ randomly initialization, $\hat{\m}$ $\gets$ $\mathbf{0}$; \\
\For{$k$ $\gets$ 1, 2, $\dots$, $\tau$ } {
Get current sparse rate via Eq.~(\ref{eq15});\\
Set the learning rate of $\hat{\m}$ via Eq.~(\ref{eq16}); \\
Get $\m$ from $\hat{\m}$ via Eq.~(\ref{eq5});\\
\For{each training step $t$ }  {
Forward propagation via ($\w \odot \m$); \\
Compute the gradient of $\hat{\m}$ via Eq.~(\ref{mask_gradient});\\
Update $\w, \hat{\m}$ using SGD optimizer; \\
}
}
\end{algorithm}
%
%

%
\begin{table}[!t]
\caption{Performance comparison of VGGNet-19 on CIFAR.}
\label{tab:cifar10}
\centering
\vspace{-0.5em}
\begin{tabular}{@{}lcccc@{}}
\toprule
Dataset  & \multicolumn{2}{c}{\;\;\;\;\; CIFAR-10 \;\;\;\;\;} & \multicolumn{2}{c}{\;\;\;\;\; CIFAR-100 \;\;\;\;\;}\\ 
\midrule
Sparse Rate & 90\% & 95\% & 90\% & 95\% \\
\cmidrule{1-1} \cmidrule(lr){2-3} \cmidrule(lr){4-5}
VGGNet-19  & 93.85   & - & 73.43  & -        \\
\cmidrule{1-1} \cmidrule(lr){2-3} \cmidrule(lr){4-5}
SET & 92.46 & 91.73 &  72.36 & 69.81 \\
SNIP & 93.63 & 93.43  & 72.84 & 71.83 \\
GraSP & 93.30 & 93.04 &72.19& 71.95 \\
SynFlow & 93.35 & 93.45 &72.24& 71.77  \\
STR & 93.73 & 93.27 & 71.93 & 71.14 \\
RigL & 93.47 & 93.35 & 71.82 & 71.53 \\
GMP & 93.59 & 93.58 & 73.10 & 72.30 \\
GraNet & 93.80 & 93.72 & 73.74 & 73.10 \\
\rowcolor[gray]{0.9} OptG  & \textbf{93.84}   & \textbf{93.79} & \textbf{73.80}  & \textbf{73.24}        \\
\bottomrule
\end{tabular}
\centering\vspace{-1mm}
\end{table}

\subsection{Quantitative Results}
\textbf{VGGNet-19}. Tab.~\ref{tab:cifar10} shows the performance of different methods for sparsifying the classic VGGNet with 19 layers on CIFAR-10/100 datasets.
Compared with the competitors, our OptG yields better accuracy under the same sparse rate on both datasets.
For instance, compared with SNIP~\cite{lee2018snip} that suffers serious performance degradation of 4.26\% when pruning 95\% parameters on CIFAR-10 (89.59\% for SNIP and 93.85\% for the baseline), the proposed OptG only lose negligible accuracy of 0.01\% (93.84\% for OptG), despite they are both built on gradient information.
On CIFAR-100, our OptG provides significantly better accuracy against other gradient-driven approaches including GrasP~\cite{wang2020picking} and RigL~\cite{evci2020rigging}, which demonstrates the superiority of optimizing the gradient-driven criteria in network sparsity.

\begin{table}[!t]
\caption{Performance comparison of ResNet-50 on ImageNet.}
\centering\vspace{-0.5em}
\label{tab:imagenet_res50}
\begin{tabular}{@{}lcccc@{}}
\toprule
Method  & Sparsity & Params &  FLOPs & Top-1 Acc.    \\ \midrule
ResNet-50 & 0.00 & 25.6M    & 4.09G   &  77.01    \\
\midrule
SNIP & 90.00 &2.56M & 409M & 67.20 \\
SET & 90.00 & 2.56M & 409M & 69.60 \\
GSM & 90.00 & 2.56M & 409M & 73.29 \\
GMP & 90.00 & 2.56M & 409M & 73.91 \\
DNW & 90.00 & 2.56M & 409M & 74.00 \\
RigL & 90.00 & 2.56M  & 960M & 73.00\\
STR & 90.55 & 2.41M & 341M & 74.01 \\
GraNet & 90.00 & 2.56M & 650M & 74.20 \\
\rowcolor[gray]{0.9} OptG & 90.00 & 2.56M & 342M & \textbf{74.55}\\
\midrule 
GMP & 95.00 & 1.28M & 204M & 70.59 \\
DNW & 95.00 & 1.28M & 204M & 68.30 \\
RigL & 95.00 & 1.28M & 490M & 70.00 \\
STR & 95.03 & 1.27M & 159M & 70.40 \\
GraNet & 95.00 & 1.28M & 490M & 72.30 \\
\rowcolor[gray]{0.9} OptG & 95.00 & 1.28M & 221M & \textbf{72.45}\\
\midrule 
RigL & 96.50 & 0.90M & 450M & 67.20 \\
STR & 96.11 & 0.99M & 127M & 67.78 \\
STR & 96.53 & 0.88M & 117M & 67.22 \\
GraNet & 96.50 & 0.90M & 368M & 70.50 \\
\rowcolor[gray]{0.9} OptG & 96.50 & 0.90M & 179M & \textbf{70.85} \\
\midrule 
GMP & 98.00 & 0.51M & 82M & 57.90 \\
DNW & 98.00 & 0.51M & 82M & 58.20 \\
STR & 97.78 & 0.57M & 80M & 62.84 \\
STR & 98.05 & 0.50M & 73M & 61.46 \\
GraNet & 98.00 & 0.51M & 199M & 64.14 \\
\rowcolor[gray]{0.9} OptG & 98.00 & 0.51M & 126M & \textbf{67.20}\\
\midrule 
GMP & 99.00 & 0.26M & 41M & 44.78 \\
STR & 98.98 & 0.26M & 47M & 51.82 \\
STR & 98.79 & 0.31M & 54M & 54.79 \\
GraNet & 99.00 & 0.26M & 123M & 58.08 \\
\rowcolor[gray]{0.9} OptG & 99.00 & 0.26M & 83M & \textbf{62.10}\\
\bottomrule
\end{tabular}
\centering\vspace{-1mm}
\end{table}

\textbf{ResNet-50}.
The comparison of compressing ResNet-50~\cite{he2016deep} between the proposed OptG and its counterparts on the large-scale ImageNet dataset is presented in Tab.~\ref{tab:imagenet_res50}. As can be seen, our OptG well surpasses its competitors across different sparse rates.
For example, in comparison with the gradient-driven approach RigL at a sparse rate of 90\%, OptG greatly reduces the FLOPs to 342M with an accuracy of 74.55\%, while RigL only reaches 73.00\% with much higher FLOPs of 960M. 
Although GraNet shows comparable accuracy at 95\% sparsity (72.30\% for GraNet and 72.45\% for OptG), its performance is built upon preserving more than 2$\times$ FLOPs than OptG (490M FLOPs for GraNet and 221M for OptG).
Further, the superiority of OptG over other methods is proportional to the sparsity level.
When the sparse rate reaches 98.00\%, all existing studies suffer severe performance degradation. In contrast, OptG presents an amazing result of 67.20\% top-1 accuracy, which well surpasses the recent advances of STR by 4.36\% and DNW by 9.00\%.
Furthermore, at an extreme sparse rate of around 99.00\%, the proposed OptG still retains a high accuracy of 62.10\%, which surpasses the second best GraNet by a large margin of 4.02\%. 
These comparison results well demonstrate the efficacy of our OptG for solving the independence paradox in compressing the large-scale ResNet.

\textbf{MobileNet-V1}. 
MobileNet-V1~\cite{howard2017mobilenets} is a lightweight network with depth-wise convolution. Thus, compared with ResNet-50, it is more different to sparsify MobileNet-V1 without performance compromise. 
Nevertheless, results on Tab.\,\ref{tab:imagenet_mobv1} show that our OptG still offers reliable performance on such a challenging task. Specifically, OptG achieves a top-1 accuracy of 70.27\% at a sparse rate of 89.00\%, which is 3.75\% higher than that of STR which suffers more parameter burden as well. A similar observation can be found when the sparse rate is around 90.00\%. Our OptG reduces the parameters to 0.41M, meanwhile it still preserves the accuracy of 66.80\%, surpassing other methods by a significant margin. Thus, OptG well demonstrates its ability to sparsify lightweight networks.

\subsection{Ablation Studies}\label{ablation}


%
\textbf{The gradual sparsity schedule.} We first perform ablation studies for our proposed sparsity schedule in Eq.~(\ref{eq15}). 
We consider the classic sparsity schedule proposed by Zhu~\emph{et al.}~\cite{zhu2017prune} for comparison, which increases the sparsity level rapidly in the early training epochs.
Tab.~\ref{tab:ablation1} lists the performance of different sparsity techniques under the same gradual sparsity schedule.
As can be observed, OptG takes the lead at both sparsity schedules, and the advantage of OptG is more obvious under our proposed schedule.
Note that applying our schedule to other methods even leads to worse performance compared with the schedule proposed by Zhu~\emph{et al.}~\cite{zhu2017prune} except for DPF that also revives weights to their original value before being sparsified.
This is reasonable for that other methods generally revive weights to 0s, which, if carried out in the latter training process, can not ensure a sufficient training. Therefore, different motivations lead to a unique schedule for OptG.

\begin{table}[!t]
\caption{Performance comparison of MobileNet-V1 on ImageNet.}
\centering\vspace{-0.5em}
\label{tab:imagenet_mobv1}
\resizebox{\columnwidth}{!}{
\begin{tabular}{@{}lcccc@{}}
\toprule
Method  & Sparsity & Params &  FLOPs & Top-1 Acc.    \\ \midrule
MobileNet-V1 & 0.00 & 4.12M    & 569M  &  71.95    \\
\midrule 
GMP & 74.11 & 1.09M & 163M & 67.70 \\
STR & 75.28 & 1.04M & 101M & 68.35 \\
STR & 79.07 & 0.88M & 81M & 66.52 \\
\rowcolor[gray]{0.9} OptG & 80.00 & 0.82M & 124M & \textbf{70.27} \\
\midrule 
GMP & 89.03 & 0.46M & 82M & 61.80 \\
STR & 85.80 & 0.60M & 55M & 64.83 \\
STR & 89.01 & 0.46M & 42M & 62.10 \\
STR & 89.62 & 0.44M & 40M & 61.51 \\
\rowcolor[gray]{0.9} OptG & 90.00 & 0.41M & 80M & \textbf{66.80} \\
\bottomrule
\end{tabular}}
\end{table}

\begin{table}[!t]
\caption{Performance comparison of ResNet-50 at 95\% sparsity on ImageNet under our proposed sparse schedule and Zhu~\emph{et al.}\cite{zhu2017prune}.}
\centering\vspace{-0.5em}
\label{tab:ablation1}
\resizebox{\columnwidth}{!}{
\begin{tabular}{@{}lcccc}
\toprule
Method  & Schedule & Top-1 Acc. & Schedule & Top-1 Acc.\\
\midrule
SET&  Zhu~\emph{et al.} & 68.40 & Ours & 66.10 \\
RigL&  Zhu~\emph{et al.} & 71.39 & Ours & 70.01 \\
DPF & Zhu~\emph{et al.} & 71.03 & Ours & 71.66 \\
\rowcolor[gray]{0.9} OptG&  Zhu~\emph{et al.} & \textbf{71.82} & Ours & \textbf{72.38}  \\
 \bottomrule
\end{tabular}}
\end{table}

\begin{figure}[!t]
\begin{center}
\includegraphics[height=0.7\linewidth]{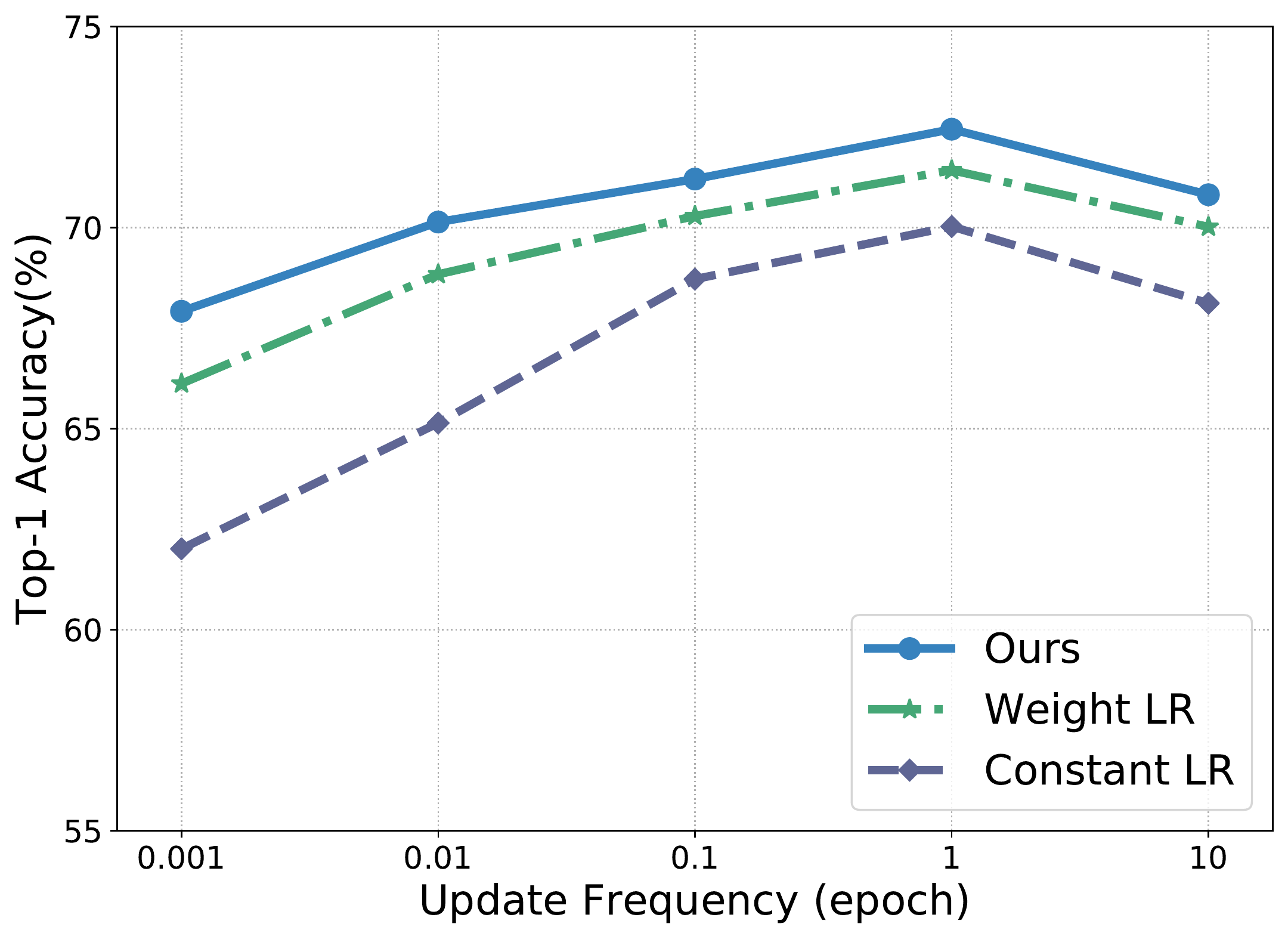}
\end{center}
\centering\vspace{0em}
\caption{\label{fig:ablation}Ablation studies for the supermask optimizer in OptG.
}
\vspace{-2em}
\end{figure}

\textbf{Supermask optimizer.} Next, we investigate the components in our proposed supermask optimizer, including the update frequency of the binary mask $\m$ and the proposed paradox-aware mask learning rate schedule. In detail, we compare our schedule with two competitors, unfolded as the same learning rate with network weights (Weight LR) and a constant learning rate of $0.1$ (Constant LR). Meanwhile, we also investigate how the update frequency of the binary mask influences the performance of OptG. Results in Fig.~\ref{fig:ablation} suggest that (1) frequently updating the binary mask,~\emph{i.e.}, prune and revive weights leads to significant performance degradation due to the unstable sparse training process and (2) our proposed paradox-aware mask learning rate schedule well surpasses the other two variants, which demonstrates the efficacy of OptG for adjusting the mask learning rate by looking at the ascending rate of sparsity so as to maximally alleviate the error-gap caused by the independence paradox.

\section{Limitation}
We further discuss the limitations of OptG, which will be our future focus.
Firstly, OptG requires more training FLOPs compared with other sparse training methods due to its specialized-designed sparsity schedule for solving the independence paradox.
Nevertheless, we state that in sparse training, less training costs rarely indicate the opportunity to lessen the training time as the speedup for the irregular sparse weight tensors on common hardware is indeed negligible.
Besides, despite the dominating of OptG at extreme sparsity levels, its performance at relatively-low sparsity rates remains to be improved.
At last, our limited hardware resources disable us to verify the efficacy of OptG beyond the convolutional neural networks. 
More results for sparsifying the popular Vision Transformer models (ViTs) are expected in our further work.

\section{Conclusion}
In this paper, we have proposed to optimize the gradient-driven criteria in network sparsity, termed OptG.
In particular, we first point out the independence paradox in previous approaches and show an effective trail to solve this paradox based on revealing the empirical success of supermask training.
Following this trail, we further propose to solve the independence paradox by interleaving the supermask training process into during-training sparsity with a revival-friendly sparsity schedule and a paradox-aware supermask optimizer.
Extensive experiments on various tasks demonstrate that our OptG can automatically obtain layer-wise sparsity burden, while achieving state-of-the-art performance at all sparsity regimes.
Our work re-emphasizes the great potential of gradient-driven pruning and we expect future advances for the gradient-driven criteria optimizer.

\section*{Acknowledgement}
This work is supported by the National Science Fund for Distinguished Young Scholars (No.62025603), the National Natural Science Foundation of China (No.U1705262, No.62072386, No.62072387, No.62072389, No.62002305, No.61772443, No.61802324 and No.61702136) and Guangdong Basic and Applied Basic Research Foundation (No.2019B1515120049).
{\small
\bibliographystyle{ieee_fullname}
\bibliography{egbib}
}

\renewcommand{\thetable}{A\arabic{table}}
\renewcommand{\thefigure}{A\arabic{figure}}
\renewcommand{\theequation}{A\arabic{equation}}
\renewcommand{\thesection}{A\arabic{section}}

\crefname{section}{Sec.}{Secs.}
\Crefname{section}{Section}{Sections}
\Crefname{table}{Table}{Tables}
\crefname{table}{Tab.}{Tabs.}

\section*{Supplmentary Material}
In this supplementary material, we first present a quantitative analysis of the independence paradox. Besides, more experimental results are presented, including the comparison with the vanilla supermasks as well as more ablation studies for OptG. Details are given in the following.
\section*{A1. Quantitative Analysis of the Independence Paradox}\label{app_paradox}
In this section, we quantitatively show the influence of the independence paradox on gradient-driven network sparsity methods.
As discussed in Sec.\;3 of the main text, the error gap caused by the independence paradox is indeed proportional to the number of removed weights at one time.
To demonstrate this, we conduct an empirical study by iteratively removing a given proportion of weights with respect to the gradient-driven criteria.
In detail, given a target sparse rate $P$, we split the sparsification process into $C$ cycles, where each cycle imposes $\frac{P}{C}$ sparsity to the current sparse networks and contains a full training schedule using the cosine annealing scheduler~\cite{loshchilov2016sgdr}.
The results displayed in Tab.\;\ref{tab:vgg} and Tab.\;\ref{tab:r50} show that once-for-all removal of weights,~\emph{i.e.}, $C=1$, acquires a stable yet relatively low performance at different total epochs.
On the contrary, iteratively removing weights brings linear probing accuracy with the total epochs, but suffers serious performance degradation when only limited epochs are given.
Particularly, when $C=100$,\;\emph{i.e.}, only $\frac{P}{100}$ percent weights are removed in each cycle, significant performance improvement is observed compared with one-for-all removal of weights at 10,000 total training epochs (93.87\% accuracy for $C=100$ and 93.26\% accuracy for $C=1$), which demonstrates that the error gap caused by the independence paradox raises proportionally to the number of removed weights at one time.
Nevertheless, the benefit of more sparsification cycles only stands with massive total epochs.
When only limited total epochs of 100 are given, setting $C=100$ only provides an accuracy of 39.81\% as the sparse weights cannot be sufficiently trained at each cycle due to the change of network typology.
In summary, the simultaneous removal of weights brings significant variation to the gradient-driven criteria due to the independence paradox, while iteratively imposing sparsification requires massive training epochs to enable sufficient training of sparse networks. 
Therefore, an efficient way to mitigate the independence paradox remains to be explored.

\begin{table}[!t]
    \caption{Top-1 accuracy (\%) for sparsifying VGGNet-19 at 95\% sparsity on CIFAR-10 with varying cycles and total epochs.}
    \centering
    \begin{tabular}{@{}c|cccc@{}}
    \toprule
    \diagbox[width=9em,trim=l]{Cycle}{Total Epochs}   & 100 & 500 & 1000 & 10000  \\
    \hline
      1 & 93.21 & 93.24 & 93.22 & 93.26\\
      5 & 92.10 & 93.35 & 93.41 & 93.42\\
      10 & 82.13 & 93.12 & 93.54 & 93.55\\
      100 & 32.11 & 61.12 &  85.13 & 93.87\\
    \bottomrule
    \end{tabular}
    \label{tab:vgg}
\end{table}

\begin{table}[!t]
    \caption{Top-1 accuracy (\%) for sparsifying ResNet-50 at 95\% sparsity on CIFAR-10 with varying cycles and total epochs.}
    \centering
    \begin{tabular}{@{}c|cccc@{}}
    \toprule
    \diagbox[width=9em,trim=l]{Cycle}{Total Epochs}   & 100 & 500 & 1000 & 10000  \\
    \hline
      1 & 92.24 & 92.20 & 92.31  & 92.33  \\
      5 & 91.17 & 92.44 & 92.94 & 93.07\\
      10 & 85.14 & 92.34 & 93.84 & 93.99  \\
      100 &39.81 & 65.14 & 90.12 & 94.85\\
    \bottomrule
    \end{tabular}
    \label{tab:r50}
\end{table}

\begin{figure}[!t]
\begin{center}
\includegraphics[height=0.65\linewidth]{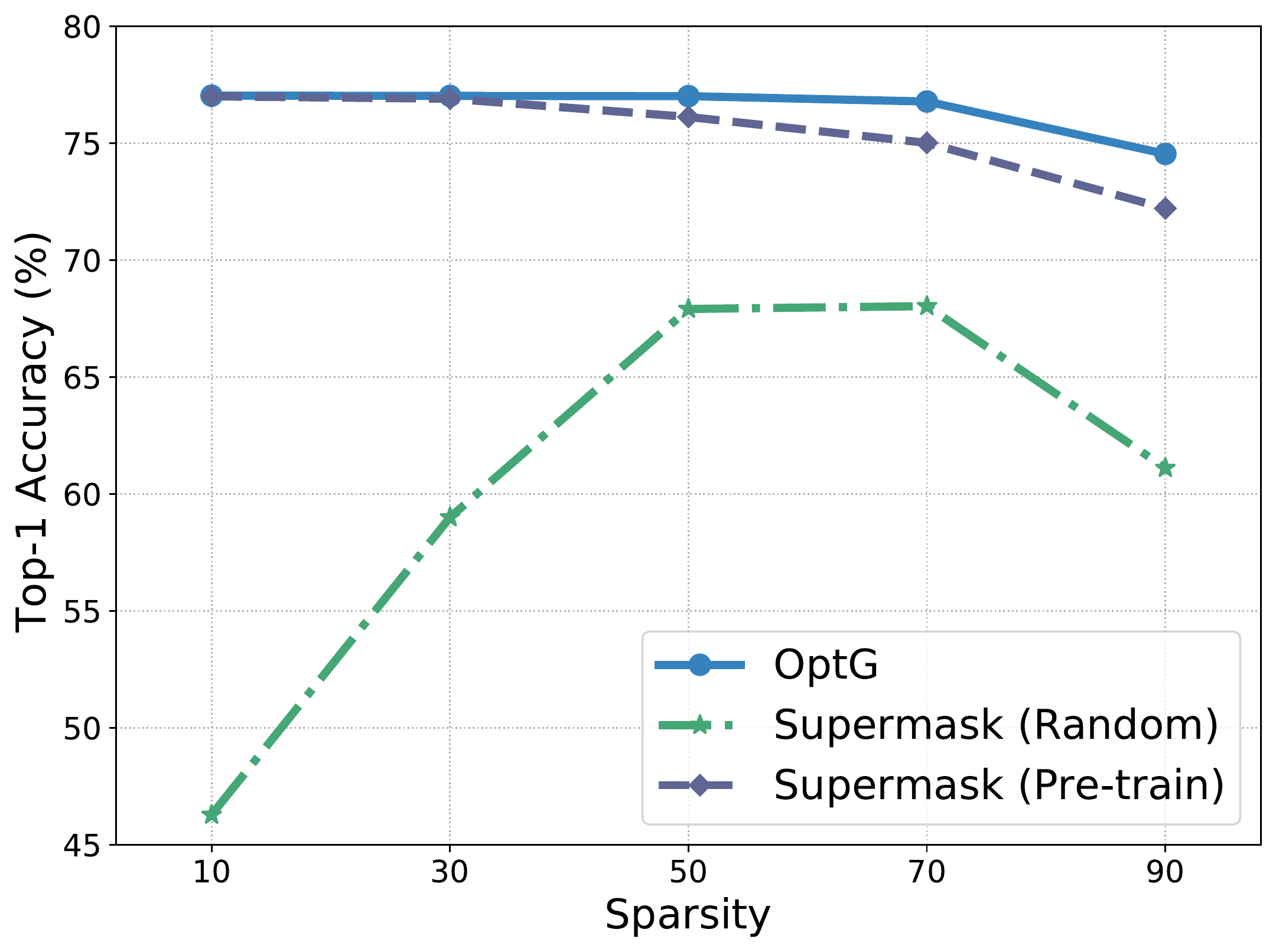}
\end{center}
\caption{\label{fig:supermask}Top-1 accuracy \emph{v.s.} sparsity with ResNet-50 on ImageNet between OptG and vanilla supermasks found from random-initialized and pre-trained weights. 
}
\end{figure}

\section*{A2. More Experimental Results}\label{app_more}
\subsection*{A2.1. Comparison with the vanilla supermasks}\label{results_supermask}
Fig.\;\ref{fig:supermask} compares the performance of OptG with the vanilla supermasks for searching subnets from randomly-initialized~\cite{ramanujan2020s} and pre-trained~\cite{zhang2021lottery} networks.
Although the vanilla supermasks partially mitigate the independence paradox as we analyze in Sec.\;3.3 of the main text, their performance degradation compared with OptG comes at the fixed weights during mask training, which prevents the training of remaining weights to comprehensively solve the independence paradox.
In addition, the supermasks that are searched on the basis of pre-trained weights surpass the randomly-initialized weights by a large margin.
This demonstrates our point in Sec.\;3.4 of the main text that the weights require sufficient training before being sparsified and revived,\;\emph{w.r.t.}, reviving random-initialized weights is meaningless.

\subsection*{A2.2. More ablation studies}\label{ablation}

\textbf{Layer-wise sparsity budgets.} We investigate the effect of layer-wise sparsity budgets by only sorting the weights inter-layer while keeping the layer-wise sparsity fixed with respect to different pre-defined budgets including Uniform, GS~\cite{han2015learning}, ERK~\cite{evci2020rigging}. As can be seen, using the global magnitude pruning (GS) and uniform budget both lead to a noticeable accuracy drop. The ERK budget can achieve better performance with our global-sorting mechanism, nevertheless, the FLOPs of ERK is much higher than the layer-wise budget automatically decided by OptG. Therefore, the efficacy of the sparsity budget chosen by OptG is also identified.

\textbf{Hyper-parameter in the sparsity schedule.}
Tab.~\ref{tab:alpha} shows the influence of the hyper-parameter $\alpha$ in the proposed sparsity schedule. 
Recall Fig.\;2 (a) in the main text, the increasing speed of sparsity is proportional to $\alpha$, and smaller $\alpha$ indicates a longer procedure for achieving the final sparsity.
A larger $\alpha$ leads to performance degradation as the rapid removal of weights brings a heavier error gap caused by the independence paradox.
Meanwhile, smaller $\alpha$ indicates delayed sparsity satisfaction and insufficient training of the final sparse networks, thus also resulting in lower accuracy.

\begin{table}[!t]
\caption{Performance of OptG for sparsifying ResNet-50 on ImageNet with different layer-wise sparsity budget.}
\label{tab:budget}

\begin{tabular}{@{}ccccc}
\toprule
Method  & Sparsity & Params &  FLOPs & Top-1 Acc.\\
\midrule
Uniform  & 90.00 &5.12M &409M & 74.12 \\
GS & 90.00 & 5.12M & 697M & 73.89   \\
ERK  & 90.00 & 5.12M  & 960M & \textbf{74.39}   \\
OptG  & 90.00 & 5.12M  & 342M & 74.28   \\
 \bottomrule
\end{tabular}
\end{table}

\begin{table}[!t]
\caption{Performance of OptG for sparsifying ResNet-50 on ImageNet with different $\alpha$ for the sparsity schedule.}
\centering
\begin{tabular}{@{}ccccc}
\toprule
$\alpha$  & Sparsity & Params &  FLOPs & Top-1 Acc.\\
\midrule
0.1  & 90.00 &5.12M & 329M & 72.01 \\
0.5 & 90.00 & 5.12M & 342M & \textbf{72.45}   \\
1  & 90.00 & 5.12M  & 361M & 72.32   \\
 \bottomrule
\end{tabular}
\label{tab:alpha}
\end{table}

\end{document}